%% file: main.tex
\documentclass[conference]{IEEEtran}

\IEEEoverridecommandlockouts
\usepackage{cite}
\usepackage{subfigure}
\usepackage{amsmath,amssymb,amsfonts}
\usepackage{algorithm,algorithmic}
\usepackage{comment}
\usepackage{graphicx}
\usepackage{textcomp}
\usepackage{xcolor}
\usepackage{caption}
\def\BibTeX{{\rm B\kern-.05em{\sc i\kern-.025em b}\kern-.08em
    T\kern-.1667em\lower.7ex\hbox{E}\kern-.125emX}}
    
\usepackage{amsthm}
\makeatletter
\def\th@plain{%
  \thm@notefont{}
  \itshape 
}
\def\th@definition{%
  \thm@notefont{}
  \normalfont 
}
\makeatother

\newtheorem{theorem}{Theorem}
\newtheorem{lemma}[theorem]{Lemma}

\newtheorem{remark}[theorem]{Remark}

\newcommand{\R}{\mathbb{R}}
\newcommand{\hypothesisset}{\mathcal{H}}
\newcommand{\cov}{\text{cov}}
\newcommand{\SP}{\text{SP}}

\begin{document}

\title{A Distributed Fair Machine Learning 
Framework with Private Demographic Data Protection\\}

\author{\IEEEauthorblockN{Hui Hu, Yijun Liu, Zhen Wang, Chao Lan}
\IEEEauthorblockA{\textit{Department of Computer Science, University of Wyoming, WY, USA} \\
Email: \{hhu1, yliu20, zwang10, clan\}@uwyo.edu}
}

\maketitle
\input{Abstract}


\input{introduction_v2}

\input{relatedwork}

\input{notation}

\input{framework}

\input{theory_v2}

\input{experiment}

\input{conclusion}

\bibliographystyle{IEEEtran}
\bibliography{reference}
\input{appendix}

\end{document}

%% file: Abstract.tex
\begin{abstract}
Fair machine learning has become a significant 
research topic with broad societal impact. 
However, most fair learning methods require 
direct access to personal demographic data, 
which is increasingly restricted to use for 
protecting user privacy (e.g. by the EU General 
Data Protection Regulation). 

In this paper, we propose a distributed fair 
learning framework for protecting the privacy 
of demographic data. We assume this data is 
privately held by a third party, which can 
communicate with the data center (responsible 
for model development) without revealing the 
demographic information. We propose a principled 
approach to design fair learning methods under 
this framework, exemplify four methods and 
show they consistently outperform their 
existing counterparts in both fairness and 
accuracy across three real-world data sets. 
We theoretically analyze the framework, and 
prove it can learn models with high fairness 
or high accuracy, with their trade-offs balanced 
by a threshold variable.
\end{abstract}

%% file: introduction_v2.tex
\section{Introduction}

It is reported that machine learning 
models are giving unfair predictions 
on minority people when being applied 
to assist consequential 
decision makings, e.g., they are biased 
against black defendants in recidivism 
prediction \cite{crime2016bias}, against female 
applicants in job hiring \cite{amazon2018bias} 
and against female employees in facial 
verification \cite{klare2012face}, etc. 
How to learn fair prediction model 
has become a pressing problem for 
government \cite{prepare2016ai}, 
industry \cite{ms2018biasface,ibm2018biasface} 
and academia \cite{nips2017fair,courtland2018bias}. 
Many fair learning methods have been developed, 
including label processing
\cite{zhang2017anti,luong2011k},   
feature processing
\cite{feldman2015certifying,zemel2013learning},   
model regularization 
\cite{dwork2012fairness,zafar2017fairness} 
and model post-processing 
\cite{hardt2016equality,fish2016confidence} 
-- some have achieved promising performance 
with very efficient trade-offs between model 
accuracy and model fairness.

We note that most fair machine learning methods 
require direct access to individuals' demographic 
data, e.g., they need individual's race information 
to mitigate racial bias.  
However, such data are increasingly restricted 
to use for protecting user privacy. 
In 2018, Europe launches a General 
Data Protection Regulation
(GDPR)\footnote{https://eugdpr.org/the-regulation/}, 
which prohibits ``processing 
of personal data revealing racial or ethnic original'' 
and allows users to request ``erasure of personal data'' 
from the data controller. Besides, 
the privacy research community has a long 
effort of hiding sensitive personal data 
from data analytics \cite{agrawal2000privacy, mohammed2011differentially}. 

We thus see fairness and privacy are running 
in a dilemma, i.e., most fair learners need 
access to demographic data while these data 
are restricted to use for privacy protection. 
Debates are arising
\cite{vzliobaite2016using,veale2017fairer}: 
should law permit the use of private demographic 
data for the sake of fair learning? is it 
technically necessary to have direct access 
to such data? Very few scientific studies 
are done to address these questions. 

In this paper, we propose a distributed 
fair machine learning framework that does 
not require direct access to demographic data. 
We assume user data are distributed 
over a data center and a third party 
-- the former holds the non-private data and is 
responsible for learning fair models; the latter 
holds the demographic data and can assist 
learning via private communications with the 
center that do not reveal user demographics. 

Based on the framework, we present a principled 
strategy to design private fair learners: 
the center first constructs a \textit{random but fair} 
hypothesis space via private communications 
with the third party; then, the center learns 
an accurate model in that space using standard methods.  
Our insight is that (i) model fairness is 
ensured by the fair hypothesis space 
and (ii) model accuracy is promised 
by random projection theory \cite{arriaga2006algorithmic,garg2002generalization}. 

Applying the strategy, we exemplify how to 
re-design four existing non-private fair 
learners into private ones, including  
fair ridge regression \cite{calders2013controlling}, 
fair logistic regression \cite{kamishima2012fairness},  
fair kernel regression \cite{perez2017fair}  
and fair PCAs \cite{samadi2018price,olfat2018convex}. 
We show the redesigned learners consistently
outperform their counterparts in both 
fairness and accuracy across three 
real-world data sets. 

Finally, we theoretically analyze the 
proposed fair machine learning framework. 
We prove upper bounds on both its model 
fairness and model accuracy, and 
show their trade-off can be balanced 
(and controlled) via a threshold 
hyper-parameter $\rho$. 

The rest of the paper is organized as follows: 
Section II introduces background and related work; 
Section III introduces notations; 
Section IV presents the proposed framework 
and exemplifies the design of four private 
fair learners; Section V presents theoretical 
analysis on the framework; Section VI 
shows experimental results and discussions; 
Section VII shows the conclusion; 
Appendix contains all proofs.

%% file: relatedwork.tex
\section{Related Work}

\subsection{Fairness Measure}

Several fairness notions have been proposed in 
the literature, such as statistical disparity 
\cite{feldman2015certifying}, 
equal odds \cite{hardt2016equality}, 
individual fairness \cite{dwork2012fairness}, 
causal fairness \cite{kusner2017counterfactual} 
and envy-free fairness \cite{balcan2018envy}. 
In this paper, we focus on statistical disparity, 
since it is most common and perhaps most refutable. 

In this paper, we propose to measure model fairness 
using covariance between prediction and demographic 
variable, as we find it extremely easy to use while 
giving very efficient accuracy-fairness trade-off. 
Similar measures have been used in the literature, 
such as mutual information \cite{kamishima2012fairness},
correlation \cite{perez2017fair} 
or independence \cite{zemel2013learning} 
between these two variables. 
But none of them provide theoretical analysis 
on the used measure.  
In this paper, we theoretically analyze the 
covariance measure; we prove low covariance 
implies low statistical disparity. 

\subsection{Fair Learning with Restricted 
Access to Demographic Data}

Several lines of studies are related to the 
restricted access of demographic data, but 
do not directly address the problem. 

A traditional fair learning method is to 
simply remove demographic feature from 
the model -- this is a natural solution to 
protect privacy. However, this approach 
does not guarantee fairness due to the 
redlining effect \cite{corbett2018measure}. 
Some studies do not use demographic 
data as a feature of the model, 
but use it in other ways during learning. 
For example, \cite{luong2011k} uses k-NN 
to detect unfair labels; they do not use 
demographic data to 
measure instance similarity, but still use 
it to measure label disparity in neighborhoods. 

Specific discussions on the restricted use of 
demographic data appears in \cite{vzliobaite2016using,veale2017fairer}; 
but there 
lacks scientific investigations or solutions. 
Recently, Kilbertus et al \cite{kilbertus2018blind} 
propose to encrypt demographic 
data before learning. This is a promising 
solution, but encryption also comes with extra 
cost of time and protocols. Our framework 
seeks another direction based on random projection; 
it is cheaper and easier to implement. 
Hashimoto et al \cite{hashimoto2018fairness} 
propose a fair learning method that 
automatically infers group membership and minimizes 
disparity across it; this method is also promising 
as it does not require access to demographic data 
at all. However, it focuses on a less common fairness 
notion called distributive justice and on-line learning. 
In contrast, we focus on the common disparity measure 
and off-line setting (although our framework is 
extendable to online setting). Besides, we hypothesize 
that one can get fairer models with even limited access 
to demographic data than with no access at all. 

Finally, studies on individual fairness do not require 
access to demographic data. For example, one can 
achieve fairness by learning a Lipschitz continuous 
prediction model \cite{dwork2012fairness}. 
Here, we focus on achieving group fairness.

%% file: notation.tex
\section{Notations}

In this section, we introduce the basic 
notations that will be used throughout the paper.
More will be introduced later. 

We will describe a random individual by a triple 
$(x, s, y)$, where $s \in \R$ is a sensitive  
demographic feature, $x \in \R^{p}$ is a 
vector of $p$ non-sensitive features 
and $y \in \R$ is the label. For example, 
when studying gender bias in hiring, 
$s$ will be an applicant's gender, $x$ 
is the non-sensitive feature vector 
(e.g. education, working hours) and 
$y$ indicates if the applicant is hired or not. 
We will index observed individuals by 
subscript, e.g.,  $(x_{i}, s_{i}, y_{i})$ 
is the $i_{th}$ individual in a (training) sample set. 

Let $f : \{ x \} \rightarrow \{ y \}$ 
be a prediction model, which does not 
take $s$ as input but can use $s$ for 
training.

%% file: framework.tex
\section{A Distributed Fair Learning Framework}

\begin{figure}[t!]
\centering
\includegraphics[width=8cm]{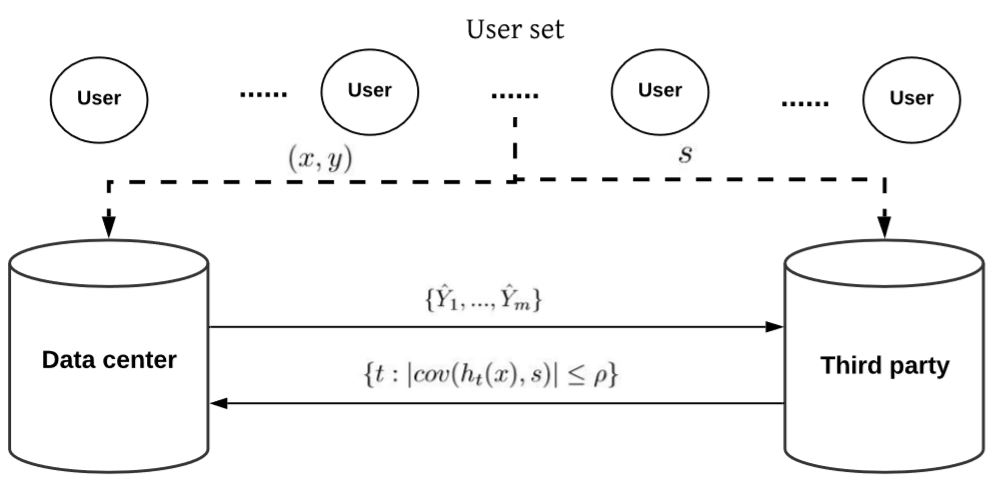}
\caption{A Distributed and Private Fair Learning Framework}
\label{fig:framework}
\end{figure}

\begin{algorithm}[t!]
 \caption{A Distributed Fair Learning Framework}
  \begin{algorithmic}[1]
 \renewcommand{\algorithmicrequire}{\textbf{Input:}}
 \renewcommand{\algorithmicensure}{\textbf{Output:}}
 \REQUIRE training set 
 $\{ (x_{i}, y_{i}) \}_{i = 1, \ldots, n}$, 
hypothesis set $\hypothesisset$, 
number of generated hypotheses $m$, 
generator variance $\sigma$, 
fairness threshold $\rho$, 
data center (DC) and third party (TP).
 \ENSURE A prediction model $f$ at DC.
\STATE DC randomly generate 
 $m$ hypotheses  $h_{1}, \ldots, h_{m} 
 \in \hypothesisset$ with each parameter 
 i.i.d. drawn from $\mathcal{N}(0,\sigma^{2})$. 
 \STATE DC applies each $h_{t}$ on 
 $\{x_{i}\}$ to get a predicted label set 
 $\hat{Y}_{t} = \{ h_{t}(x_{1}), h_{t}(x_{2}), 
 \ldots, h_{t}(x_{n})\}$.
 \STATE DC sends $\hat{Y}_{1}, \hat{Y}_{2}, 
 \ldots, \hat{Y}_{m}$ to TP. 
 \STATE TP estimates $\cov(h_{t}(x), s)$ 
 from $\hat{Y}_{t}$ and $\{ s_{i} \}$ for 
 each $t$, and returns $t$ to DC if $
 |\cov( h_{t}(x), s )| \leq \rho$.
 \STATE DC receives a set of returned 
 indices $r_{1}, r_{2}, \ldots, r_{k}$, 
 and trains a prediction model $f$ 
 on $\{ (x_{i}, y_{i}) \}$ assuming that 
 \begin{equation}
 \label{eq:finalhypothesis}
     f = \alpha_{1} h_{r_{1}} + 
     \alpha_{2} h_{r_{2}} + \ldots 
     + \alpha_{k} h_{r_{k}},  
 \end{equation}
 where $\vec{\alpha}$ = $[\alpha_{1}, \ldots, \alpha_{k}]^{T}$ 
 is unknown coefficient to learn. 
 \end{algorithmic} 
 \label{alg:dfl}
 \end{algorithm}
 
In this section, we present the proposed 
fair learning framework and exemplify how to 
design private fair learner with it. 

We assume a scenario in Figure
\ref{fig:framework}:   
there is a data center and a third party, 
over which a training set  
$\{(x_{i}, s_{i}, y_{i})\}_{i=1,\ldots,n}$ 
is distributed.  
The center has $\{(x_{i}, y_{i})\}$ 
and focuses on learning fair model $f$; 
the party has $\{s_{i}\}$ and can assist 
learning via private communications with 
the center that reveal no $s$. 

Our strategy to design fair learner is 
shown in Algorithm \ref{alg:dfl}. 
It has two phases: (i) steps 1 to 4 construct 
a random and fair hypothesis space spanned 
by $h_{r_{1}}, \ldots, h_{r_{k}}$; 
(ii) step 5 learns an accurate model 
in that space. 

Specifically, the center first generates 
$m$ random hypotheses from Gaussian distributions 
(step 1), gets their predictions on the training set 
(step 2) and sends these predictions to the 
third party (step 3). The party estimates 
correlation between its demographic data 
and each hypothesis's prediction;  
if a correlation is small enough, the center 
will be informed that the corresponding 
hypothesis is fair (step 4). Finally, 
the center will learn an accurate model 
spanned by all fair random hypotheses 
-- the model will be both fair and accurate. 
Note that, throughout the process, demographic 
data is not revealed to the center and hence 
its privacy is protected. 

Next, we exemplify how to apply 
Algorithm \ref{alg:dfl} to redesign four existing 
non-private fair learners into private ones. 
These four learners are chosen as they 
are fundamental and cover different settings, 
namely, linear vs non-linear, regression 
vs classification, and predictive learning 
vs feature learning. 
More sophisticated learners may 
be designed in similar ways. 

For ease of discussion, 
we will write $X$ = $[x_{1}, \ldots, 
x_{n}]^{T}$ as a sample matrix, $Y$ = $[y_{1}, 
\ldots, y_{n}]^{T}$ as the associated label 
vector and $H = [h_{r_{1}}, \ldots, h_{r_{k}}]$ 
as a matrix of returned hypotheses. 
Since $\vec{\alpha} = [\alpha_{1}, \ldots,
\alpha_{k}]^{T}$, we will write 
\begin{equation}
f = \sum_{t=1}^{k}
\alpha_{t} h_{r_{t}} = H \vec{\alpha}.    
\end{equation} 

\input{algorithm}

%% file: algorithm.tex
\subsection{Distributed Fair Ridge 
Regression (DFRR)}
\label{subsec:fairregression}

Calders et al \cite{calders2013controlling} 
develop a fair ridge regression (FRR). 
It minimizes squared loss on training sample, 
while additionally penalizing prediction disparity 
across demographic groups. Let $I_{1}, I_{2}$ be the 
index sets of two demographic groups (e.g. female and male) 
respectively. Their objective function is
\begin{equation*}
J_{FRR}(f) = {\sum}_{i=1}^{n} (f(x_{i}) - y_{i})^{2} 
+ \lambda \cdot \text{MD}(f),     
\end{equation*}
where $\text{MD}(f) = \frac{1}{|I_{1}|} \sum_{i \in I_{1}} 
f(x_{i}) - \frac{1}{|I_{2}|} \sum_{i \in I_{2}} f(x_{i})$ 
is the prediction disparity. 
We see $\min J(f)$ requires simultaneous access 
to $(x,y)$ and $s$; thus this method cannot be directly 
applied in our private learning framework. 

We propose a distributed fair ridge regression 
(DFRR) based on Algorithm \ref{alg:dfl}. 
Our objective function is 
\begin{align}
\begin{split}
 & J_{DFRR}(f) = 
\sum_{i=1}^{n} (f(x_{i})-y_i)^{2}
+ \lambda ||f||^{2},\\ 
&\quad\ = \sum_{i=1}^{n} \left(
\sum_{t=1}^{k} \alpha_{t} h_{r_{t}}(x_{i}) 
-y_i\right)^{2} 
+ \lambda \left\| \sum_{t=1}^{k} \alpha_{t} 
h_{r_{t}} \right\|^{2}.   
\end{split}
\end{align}
Minimizing the above objective for $\alpha_{t}$'s 
gives 
\begin{equation}
\label{eq:dfrr}
\vec{\alpha} = (H^T X^T X H + \lambda I)^{-1} (H^{T} X^{T} Y).
\end{equation}

The general argument for (\ref{eq:dfrr}) is 
that  we can first solve for $H \vec{\alpha}$ 
(by standard method such as least square), 
and then solve for $\vec{\alpha}$. This 
argument will be repeatedly used in the sequel. 

\subsection{Distributed Fair Kernel Ridge 
Regression (DFKRR)}
\label{subsec:fairkernel}

Perez-Suay et al\cite{perez2017fair} develop
a fair kernel ridge regression (FKRR). It minimizes 
squared loss in RKHS while additionally 
penalizing the correlation between prediction and 
demographic feature. Its objective function is 
\begin{equation*}
J_{FKRR}(f) 
= {\sum}_{i=1}^{n} (f(\phi(x_{i})) - y_{i})^{2} 
+ \lambda \Omega(f) + \mu I(f; s),      
\end{equation*}
where $I(f; s) = \sum_{i=1}^{n} 
(\bar{f}(x_{i}) \cdot \bar{s}_{i})$ is 
the correlation between prediction and 
demographic and $\bar{f}$ and $\bar{s}$ 
are centered variables. 
This method also needs 
simultaneous access to $(x,y)$ and $s$. 

We present a distributed fair 
kernel regression (DFKRR) method 
based on Algorithm \ref{alg:dfl}.
Our high-level objective is 
\begin{equation}
J_{DFKRR}(f) 
= \sum_{i=1}^{n} (f(\phi(x_{i}))-y_i)^{2} 
+ \lambda ||f||^{2}.    
\end{equation}
Unlike the standard assumption that $f$ is 
expressed by $\phi(x_{i})$'s, we first assume 
$f$ is expressed by $h_{r_{t}}$'s as in (1) 
and each $h_{r_{t}}$ is linearly 
expressed by $\phi(x_{i})$'s, i.e., 
\begin{equation}
h_{r_{t}} = \sum_{i=1}^{n} c_{ti} \phi(x_{i}),  
\end{equation}
where $c_{ti}$'s are random coefficients 
associated with $h_{r_{t}}$. A similar argument 
has been used \cite{grandvalet2003scaling}.

Based on (5), we can generate a random hypothesis 
(and its predicted label set) by randomly generating 
a set of associated coefficients. 
Note the coefficients $c_{t_{i}}$'s are known and 
$\alpha_{t}$'s are unknown. Minimizing $J(f)$ gives 
\begin{equation}
\vec{\alpha} = 
 [C^T(K+\lambda I)^{T}(K+\lambda I)C]^{-1}C^{T}
 (K+\lambda I)^{T}Y.
\end{equation}
where ${K}$ is the Gram matrix and 
$C$ is an $n$-by-$k$ matrix with 
$c_{ti}$ being its element at 
the $i_{th}$ row and $t_{th}$ column.

\subsection{Distributed Fair Logistic 
Regression (DFGR)} 
\label{subsec:fairlogit}

Kamishima et al \cite{kamishima2012fairness} 
developed a fair logistic regression (FGR). 
It maximizes the likelihood of label while 
additionally penalizing mutual information 
between model prediction and demographic 
feature. Its objective function is 
\begin{equation*}
J_{FGR}(f) = - {\sum}_{i=1}^{n} \ln p(y_{i} \mid 
x_{i}, s_{i}, f) + \frac{\lambda}{2} ||f||^{2} + \text{R}(f), 
\end{equation*}
where $\text{R}(f) = \text{E}\, p(f(x),s) \ln 
\frac{p(f(x),s)}{p(f(x)) p(s)}$ measures the 
mutual information and can be estimated from data. 
This method also 
requires simultaneous access to $(x,y)$ and $s$. 

We propose a distributed fair logistic regression 
(DFGR) based on Algorithm \ref{alg:dfl}. 
Our high-level objective function is 
\begin{equation}
\label{eq:dfgr}
J_{DFGR}(f) = - \sum_{i=1}^{n} \ln p(y_{i} 
\mid x_{i}, f) + \lambda ||f||^{2}, 
\end{equation}
where $p(y_{i} | x_{i}, f)$ is constructed 
in the same way as logistic regression, 
with an additional assumption  
$f$ has the form (1). 

Minimizing (\ref{eq:dfgr}) by Newton's 
method, we can update
\begin{equation}
\vec{\alpha} = \vec{\alpha} 
- (J^{''}(f))^{-1} (J^{'}(f)),     
\end{equation}
where
\begin{equation}
J^{'}(f)=H^{T}X^{T}(Y-\vec{p})+2\lambda H^{T}H 
\vec{\alpha},     
\end{equation}
and
\begin{equation}
J^{''}(f)=H^{T}X^{T}MXH+2\lambda H^{T}H,  
\end{equation}
with 
${\vec{p}=[p(f(x_1)=1|x_1;f),
\ldots,p(f(x_n)=1|x_n;f)]^{T}}$ and 
diagonal matrix ${M}$ with 
$M_{ii} = p(f(x_i)=1|x_i;f) 
\cdot p(f(x_i)=0|x_i;f)$ -- both 
are standard quantities in logistic 
regression.

\subsection{Distributed Fair PCA}
\label{subsec:fairpca}

Samadi et al \cite{samadi2018price} 
develop a fair PCA that minimizes 
reconstruction error while equalizing 
this error across demographic groups. 
Let $X_{1} \in \R^{n_{1} \times p}$ be the 
sample matrix of $n_{1}$ instances in 
one group, $X_{2} \in \R^{n_{2} \times p}$ 
be the sample matrix of $n_{2}$ instances 
in another group, and $V \in \R^{p \times q}$  
be the projection matrix. Their 
objective (to minimize) is 
\begin{equation*}
\max \left\{ \frac{1}{n_{1}} 
loss(X_{1}, X_{1} V V^{T}),\, 
\frac{1}{n_{2}} loss(X_{2}, X_{2} V V^{T}) \right\}.  
\end{equation*}
where $loss$ measures reconstruction error. 
Authors show the optimal $V$ gives equal  
reconstruction errors across groups. 

Matt Olfat et al\cite{olfat2018convex} 
propose another fair PCA method that minimizes 
prediction disparity in the projected space, 
i.e., 
\begin{equation*}
\min_{w, V}\,\sup_t\, | p[w^{T}V^{T}x 
\leq t|s=1]-p[w^{T}V^{T}x \leq t|s=0] |,   
\end{equation*}
where $w$ is the prediction model and $V$ 
is the project matrix. 

Note that both methods need 
access to $(x,y)$ and $s$. 

We propose a distributed fair PCA (DFPCA) 
method based on Algorithm \ref{alg:dfl}. 
Let $v$ be a projection vector. 
Our optimization probelm is the same as PCA, i.e., \begin{equation}
\label{pr:10}
\max_{v} v^T \Sigma_{x} v,  
\quad \text{s.t.} ||v||=1. 
\end{equation} 
where $\Sigma_{x}$ is the covariance matrix. 
Our additional assumption is that $v$ 
is linearly expressed by fair random 
hypotheses, i.e., 
\begin{equation}
v = \alpha_1 h_{r_{1}} + \ldots 
+ \alpha_k h_{r_{k}}  = H \vec{\alpha}. 
\end{equation} 
Solving problem (\ref{pr:10}) 
for $\vec{\alpha}$ gives 
\begin{equation} 
H^T \Sigma_{x} H \vec{\alpha} 
= \lambda H^{T} H \vec{\alpha},  
\end{equation}
which implies $\vec{\alpha}$ is the 
leading (generalized) eigenvector.

%% file: theory_v2.tex
\section{Theoretical Analysis}
\label{sec:theory}

Here we present the
theoretical properties of 
Algorithm \ref{alg:dfl}. 

\subsection{Preliminaries}

Let $(x,s)$ be a random instance. 
We say a hypothesis $f$ is {$\rho$-fair} 
with respect to $s$ if $|\cov[f(x), s]| \leq \rho$. Note it means, in Algorithm \ref{alg:dfl}, 
all returned hypotheses $h_{r_{1}}, \ldots, h_{r_{k}}$ are $\rho$-fair. 

We will show $\rho$-fairness implies 
a popular fairness measure called 
\textit{statistical parity} (SP) \cite{mcnamara2017provably}, defined as
\begin{equation}
\SP (f) = |p(f(x)=1|s=1)-p(f(x)=1|s=0)|.  
\end{equation}

To establish the implication, we will employ 
the following generalized covariance inequality 
\cite[Theorem 2]{matula2003some}. 
\begin{lemma}
\label{thm:lm1}
Let $X, Y$ be two positively or negatively 
quadrant dependent random integers.  
Let $F_{X,Y}(x,y)$ be their joint CDF 
and $F_{X}(x)$, $F_{Y}(y)$ be their 
marginal CDF's respectively. Let 
\begin{equation}
\cov_H(X,Y) = \int_{-\infty}^{\infty}
\int_{-\infty}^{\infty} \Delta F_{X,Y}(x,y) dxdy,   
\end{equation}
be their Hoeffding covariance, where 
\begin{equation}
\Delta F_{X,Y}(x,y) = F_{X,Y}(x,y) 
- F_{X}(x) F_{Y}(y).     
\end{equation}
If $\cov_H(X,Y)$ is bounded, then 
\begin{equation}
{\sup}_{x,y} |\Delta F_{X,Y}(x,y)| 
\leq |\cov_H(X,Y)|.
\end{equation}
\end{lemma}

In the following, we will first 
present theoretical properties on model 
fairness and then on model error. Note 
that all results are presented in the 
context of Algorithm \ref{alg:dfl}. 

\subsection{Theoretical Properties on 
Model Fairness}
Our first result shows that $\rho$-fair 
implies statistical parity.

\begin{lemma}
\label{thm_sp}
If $f(x)$ and $s$ are positively 
or negatively quadrant dependent\footnote{
Quadrant dependence is a common 
assumption e.g., \cite{denuit2004nonparametric,racine2015mixed}.  
Later we will show empirical evidence 
that our assumption holds in most cases.}, 
and if $f$ is $\rho$-fair with respect to $s$, 
then $\SP(f) \leq  \rho / s_{0} s_{1}$,    
where $s_{0}$ = $p(s=0)$ and 
$s_{1}$ = $p(s=1)$.
\end{lemma}

Our second result suggests that a hypothesis 
spanned by fair hypotheses remains fair -- 
this is the 
insight that motivates the study. 
More specifically, in (1), we show 
that $f$ is $\rho$-fair because  
it is spanned by $\rho$-fair hypotheses 
$h_{r_{1}}, \ldots, h_{r_{k}}$. 

\begin{lemma}
\label{lemma_fairF}
In (\ref{eq:finalhypothesis}), 
$f$ is $(\sqrt{k}\, ||\vec{\alpha}||
\rho)$-fair w.r.t. $s$. 
\end{lemma}

Combining the above result, 
we immediately have 

\begin{theorem}
\label{thm:finalfair}
In Algorithm \ref{alg:dfl}, 
if $f(x)$ and $s$ are positively 
or negatively quadrant dependent, 
then $\SP(f) \leq \sqrt{k} ||\vec{\alpha}|| 
\rho / s_{0} s_{1}$.
\end{theorem}

This theorem implies one can obtain a 
fair model through several paths. 
First, we can choose a small threshold 
$\rho$, which will reduce prediction 
disparity at a rate of $O(\rho)$. Another 
way is to choose a small $k$ but it does 
not seem very efficient as (i) it has 
a lower reduction rate $O(\sqrt{k})$ 
and (ii) it can be implied by choosing 
a small $\rho$ (thus returning fewer 
hypotheses). 

One may also choose a small $||\vec{\alpha}||$. 
In our proposed methods, this is 
done indirectly via regularizing 
$||f||$. In experiments, we observe this 
is more effective than directly regularizing $\vec{\alpha}$. 

Finally, we see a model may be more fair 
if the demographic distribution is more 
balanced, i.e., the upper bound of $\SP(f)$ 
is minimized when $s_{0}$ = $s_{1}$ = 0.5. 
However, such distribution is typically 
formed by nature and cannot be easily 
modified. 

Our following result gives more insight 
on the number of returned hypotheses $k$, 
and suggests it shall not be too small. 

\begin{lemma}
\label{lemma_k}
Let $h$ be a random hypothesis. Then 
\begin{equation}
\text{E}\, [k] \geq m \cdot 
(1 - \text{E}\, [\cov(h(x),s)]\, / \rho),   
\end{equation}
where both expectations are taken over 
the randomness of $h$, and the covariance 
is defined over the randomness of $(x,s)$. 
Further, if $h$ is linear and generated 
from $\mathcal{N}(0, \sigma^{2} I)$, then 
\begin{equation}
\text{E}\, [k] \geq m 
\cdot ( 1 - \sigma^2 
||\vec{\cov}(x,s)||^2 / \rho^2),
\end{equation}
where $\vec{\cov}(x,s)$=$\sum_{j=1}^{p} 
\cov(x_{j},s)$ and $x_{j}$ is 
$j_{th}$ entry of $x$.
\end{lemma}

Lemma \ref{lemma_k} implies that $E[k]$ 
increases as $\rho$ increases, and the 
rate can be larger if $h$ is linear; 
when $\rho$ approaches infinity, $E[k] 
\geq m$ which means all hypotheses will 
be returned. The lemma also implies 
that smaller $\sigma$ implies larger $k$. 

\subsection{Theoretical Properties on 
Model Generalization Error}

To derive an error bound for the algorithm, 
our backbone technique is the random projection 
theory \cite{garg2002generalization}. 
It states that data distance 
is likely to be preserved in a randomly projected 
space and thus a model's prediction error 
(dependent on such distance) is also likely to be preserved. 

To apply the theory, we assume $f, h$ are linear 
and interpret the returned hypotheses as basis of a 
randomly projected space, i.e., $h_{r_{k}}(x)$ is 
the $k_{th}$ feature of $x$ in the projected space. 

We also assume Step 4 applies a  
\textit{soft} threshold policy.   
Let $h_{*}$ be a hypothesis 
satisfying $\cov(h_{*}(x),s)$ = 0.   
The soft policy will return $t$ 
of any $h_{t}$ with probability 
$\mathcal{N}(\vec{h}_{*}(x), 
\sigma^{2}_{2} I)$, 
where $\vec{h}_{*}(x)$ = $[h_{*}(x_{1}), 
\ldots, h_{*}(x_{n})]^{T}$ and 
$\sigma_{2}$ is constant. 
As such, each returned hypothesis $h_{r}$ 
in (\ref{eq:finalhypothesis}) is first drawn 
from a zero-mean Gaussian (Step 1) and then 
selected by a $\vec{h}_{*}$-mean Gaussian 
(Step 4). Therefore, we can say each $h_{r}$ 
in (\ref{eq:finalhypothesis}) is generated 
from a Gaussian centered at $\vec{h}_{*}$. 
Without loss of generality, we assume 
this Gaussian has a unit variance. 

Our first result extends the data distortion bound 
in \cite{arriaga2006algorithmic} from 
zero-mean Gaussian to non-zero mean Gaussian.  

\begin{lemma}
\label{lemma_distortion}
Let $x$ be any point and $H = [h_{r1}, \ldots, 
h_{r_{k}}]$ be a projection matrix with each 
projection vector $h_{r_{t}}$ taken
from a normal distribution 
$\mathcal{N}(h_{*}, I)$. 
Let $\tilde{x} = \frac{1}{\sqrt{k}}(H^{T} x)$ 
be the projection of $x$ by $H$. 
We have for $0 \leq c < 1$, 
\begin{equation}
\Pr\{ |\,||\tilde{x}||^{2} - ||x||^{2}\,|
\geq c ||x||^{2}\} \leq g(x) \cdot e^{- 
\frac{c^{2} k }{8}},   
\end{equation}
where $g(x) = 
e^{(ck \langle h_{*}, x\rangle^{2})
/(4 - 2c)} + e^{- c k \langle h_{*},
x\rangle^{2}) / (2 + 2c)}$. 
\end{lemma}

Compared to the original bound, our new 
bound has an additional term $g(x)$. 
It is smaller when $||h_{*}||$ is smaller; 
if $h_{*}$ = $0$, then $g(x)$ = $2$ 
and we recover the original bound. 

Based on Lemma \ref{lemma_distortion}, 
we derive the following error bound. 

\begin{theorem}
\label{thm_error}
Suppose Algorithm \ref{alg:dfl} 
adopts the soft threshold policy. 
Let $er(f)$ and $\hat{er}(f)$ 
be the expected and empirical 
error of $f$ respectively. 
If $f$ is linear and 
$||f|| = ||x|| = 1$, then 
with probability at least $1-4\delta$, 
\begin{equation}
er(h) \leq \hat{er}(h) + 
 T + \frac{4}{n\delta} 
\sum_{i=1}^{n} g(x_{i}) e^{\frac{- k \langle 
f, x_{i} \rangle^{2}}{8(2+||\langle f, x_{i} 
\rangle||)^{2}}}, 
\end{equation}
where  $T = 2\sqrt{
 [ (k+1) \log (en/(k+1)) 
 + \log 1/\delta ] / n }$ and 
\begin{equation}
g(x_{i}) = e^{
(c_{i} k \langle h_{*}, 
x_{i} \rangle^{2})/(4 - 2c_{i})} 
+ e^{- (c_{i} k \langle h_{*}, 
x_{i} \rangle^{2}) / (2 + 2c_{i})}     
\end{equation} 
with $c_{i} = |\langle f, x_{i} \rangle| 
/ (2 + |\langle f, x_{i} \rangle|)$.
\end{theorem}

An important parameter in the 
error bound is $k$. 
To facilitate discussion, we can 
loosen the bound and have 

\begin{remark}
In Theorem \ref{thm_error}, if 
\begin{equation}
\left(\langle h_{*}, x_{i} \rangle^{2} 
- 1/4 \right)  
(||\langle f, x_{i} \rangle|| - 2 )^{2} + 1 \leq 0, 
\end{equation}
then there exist positive constants 
$c_{1}$ and $c_{2}$ such that 
\begin{equation}
er(h) \leq \hat{er}(h) 
+ c_{1} + O(e^{-c_{2} k}). 
\end{equation}
\end{remark}

We see error bound decreases 
exponentially as $k$ increases, 
suggesting one choose large $k$ 
to get accurate models. 
Note this is opposite to 
Theorem \ref{thm:finalfair},
which suggests choosing small 
$k$ to get fair models. 
So we see a trade-off 
between accuracy and fairness 
is established (and controlled) 
via parameter $k$. In practice, 
we can adjust $k$ by adjusting 
the threshold $\rho$.

%% file: experiment.tex
\section{Experiment}

In this section, we evaluate the 
proposed distributed and private fair 
learning methods on three real-world data sets, 
and compared them with their existing 
non-private counterparts. 
To facilitate reproduction of the present results, 
we published our experimented 
data sets and random index sets at\footnote{https://uwyomachinelearning.github.io/} 
and the codes of our implemented methods at 
\footnote{https://github.com/HuiHu1/Distributed-Private-Fair-Learning}. 

\subsection{Data Preparation}

We experimented on three public data sets 
commonly used for evaluating algorithm 
fairness: the Community Crime data 
set\footnote{https://archive.ics.uci.edu/ml/datasets/default+of+credit+card+clients}, 
the COMPAS data 
set\footnote{https://www.kaggle.com/danofer/compass} 
and the Credit Card data set\footnote{https://archive.ics.uci.edu/ml/datasets/default+of+credit+card+clients}. 

The Community set contains 1993 communities described 
by 101 features; community crime rate is the label;  
we treated a community as minority if its 
fraction of African-American residents is 
greater than 0.5. 
The COMPAS set contains 18317 records 
described by 40 features; risk of recidivism is 
the label; we removed incomplete data and 
ended up having 16000 records and 15 features; 
similar to \cite{chouldechova2017fair}, 
we treated race as the sensitive feature. 
The Credit set contains 
30000 users described by 23 features; 
default payment is the label;  
similar to \cite{samadi2018price},
we selected education degree as the 
sensitive feature. 

\subsection{Experiment Design}

On each set, we randomly chose $75\%$ 
instances for training and used the 
rest for testing. We evaluated each 
method for 50 random trials and reported 
its averaged performance. 

We compared each proposed distributed 
private fair learner with its existing 
non-distributed non-private counterpart, i.e.,
\begin{itemize}
\item Distributed Fair Ridge Regression 
(DFRR) vs Fair Ridge Regression (FRR)\cite{calders2013controlling}
\item Distributed Fair Logistic Regression 
(DFGR) vs Fair Logistic Regression 
(FGR)\cite{kamishima2012fairness} 
\item Distributed Fair Kernel Regression 
(DFKRR) vs Fair Kernel Regression 
(FKRR)\cite{perez2017fair} 
\item Distributed Fair PCA (DFPCA) vs 
two Fair PCA (FPCA) methods \cite{samadi2018price,olfat2018convex}
\end{itemize}
We also compared with a popular 
fair learner LFP\cite{zemel2013learning}.
For competing methods, we use their 
default hyper-parameters (or, grid-search 
from the default candidate values) 
identified in previous studies. In experiment, 
we observe these configurations generally 
achieve best performance. For our proposed 
DFGR, learning rate was set to 0.001. 

We used five evaluation metrics: 
statistical parity (SP) \cite{mcnamara2017provably}, normed disparate (ND) \cite{mcnamara2017provably}, 
classifier error, error parity and error disparate. 
Let $er(f | s = 1)$, $er(f | s = 0)$ be the 
classifier errors in two demographic groups 
respectively. We define 
\begin{equation}
 \text{Error Parity}(f) = | er(f | s = 1) 
 - er(f | s = 0) |. 
\end{equation}
and 
\begin{equation}
\text{Error Disparate}(f) = \left| 
\frac{er(f | s = 1)}{er(f | s = 0)} - 1 \right|.     
\end{equation}
Small SP, ND, EP and ED implies fair models; 
small classifier error implies accurate model. 

\subsection{Results and Discussions}

Our experimental results on the three data sets 
are presented in Tables \ref{tab1}, \ref{tab2} 
and \ref{tab3} respectively. 
For the proposed learners, we set $\rho$ to 
0.01, 0.1 and 0.25 on the three sets respectively. 
Our discussions will focus on Table \ref{tab1}. 

First, we observe the proposed distributed 
and private fair learners consistently outperform 
their non-private counterparts. Take ridge 
regression as an example, DFRR not only achieves 
much lower SP than FRR (0.05 vs 0.31), but 
also achieves lower classifier error (0.106 vs 0.110) and error parity (0.17 vs 0.23). 
Another example is PCA, where DFPCA achieves 
lower SP than FPCA's (0.03 vs 0.08), lower 
classifier error (0.14 vs 0.15) and lower 
error parity (0.15 vs 0.19). Similar observations 
can be found on other two data sets. 
This implies two things: (1) the proposed 
distributed fair learning framework is effective; 
(2) the proposed private fair learners can achieve 
more efficient trade-off between fairness and 
accuracy than the state-of-the-art 
non-private counterparts.  

Our second observation is that the performance gap 
between private and non-private fair learners is 
larger for linear models (ridge regression and PCA) 
compared with nonlinear models (logistic and kernel). 
This is partly consistent with the theoretical 
guarantees we proved for linear models. As to why 
our framework gives less improvement on non-linear
models, we do not have a principled hypothesis at 
the moment. 

Finally, we see previous fair PCA methods do 
not improve fairness in classification tasks. 
Comparatively, our proposed
distributed and private
fair PCA significantly reduces 
SP and classifier error, making
itself competitive for fair 
classification. 

\begin{table*}[t!]
\def\arraystretch{1.3}
\small 
\label{tab1}
\caption{Classification Performance on the Community 
Crime Data Set }
\vspace{-5pt}
\begin{center}
\begin{tabular}{l||c|c||c|c|c}
\hline
\textbf{Method} & \bf Statistical 
Parity & \bf Normed Disparate & 
\bf Classifier Error 
& \bf Error Parity & \bf Error Disparate \\ \hline\hline
FRR\cite{calders2013controlling} & .3062$\pm$.0452 
& .2457$\pm$.0128 & .1102$\pm$.0128 
& .2260&.7321 \\ \hline 
DFRR&.0466$\pm$.0117 & .1691$\pm$.1081 & 
.1064$\pm$.0092 & .1727&.6866 \\ \hline
FKRR \cite{perez2017fair} &.0968$\pm$.0722 & .1274$\pm$.0105 
& .1208$\pm$.0054 & .1250 &.2515\\ \hline
DFKRR &.0695$\pm$.0181 & .1060$\pm$.0081 & .1216$\pm$.0143 & .1152 &.2510 \\ \hline
FGR \cite{kamishima2012fairness}  &.0898$\pm$.0971 
& .1154$\pm$.0308 &.1166$\pm$.0189 
& .1424 &.5723\\ \hline 
DFGR &.0650$\pm$.0198 & .1097$\pm$.0872 &.1202$\pm$.0690 
&.1212 &.5190\\ \hline
FPCA1 \cite{samadi2018price} & .0859$\pm$.0479 & .3546$\pm$.0225 & .1731$\pm$.0089 & .1895 &.5557\\ \hline 
FPCA2\cite{olfat2018convex} &.0755$\pm$.0293  & .3319$\pm$.0186 & .1476$\pm$.0122 & .1851 &.6091\\ \hline 
DFPCA & .0289$\pm$.0502& .2263$\pm$.0306 &.1351$\pm$.0111
&.1502 &.6507\\ \hline 
LFR\cite{zemel2013learning}  & .0738$\pm$.0377 & .2240$\pm$.0194 &.1264$\pm$.0068 &.1319 &.5431\\ \hline \end{tabular}
\end{center}
\label{tab1}
\end{table*}


\begin{table*}[t!]
\def\arraystretch{1.3}
\small
\label{tab2}
\caption{Classification Performance 
on the COMPAS Data Set}
\vspace{-5pt}
\begin{center}
\begin{tabular}{l||c|c|c||c|c}\hline
\textbf{Method} & \bf Statistical 
Parity & \bf Normed Disparate & 
\bf Classifier Error
& \bf Error Parity & \bf Error Disparate\\ \hline\hline
FRR\cite{calders2013controlling} & .0515$\pm$.0042 & .2361$\pm$.0414
& .2276$\pm$.0040  & .0317 &.1081\\ \hline 
DFRR &.0078$\pm$.0041 &.1758$\pm$.0987&.2302$\pm$.0045&.0139 &.0543\\ \hline
FKRR\cite{perez2017fair} &.0041$\pm$.0013&.1194$\pm$.0237&.2190$\pm$.0089&.0027&.0122\\ \hline
DFKRR &.0034$\pm$.0015&.1147$\pm$.0688&.2152$\pm$.0093&.0017&.0078\\ \hline
FGR\cite{kamishima2012fairness}&.0408$\pm$.0162&.2842$\pm$.0319&.2428$\pm$.0917&.0222 &.0865\\ \hline 
DFGR&.0374$\pm$.0645 &.1852$\pm$.0973&.2617$\pm$.0509&.0104 &.0385\\ \hline 
FPCA1\cite{samadi2018price}&.2806$\pm$.0182&.3028$\pm$.0232&.3204$\pm$.1032 &.0429&.1190\\ \hline
FPCA2\cite{olfat2018convex}&.1719$\pm$.0317&.2901$\pm$.1027&.2390$\pm$.0278&.0394&.1472\\ \hline 
DFPCA&.0081$\pm$.0046 &.2019$\pm$.1011&.2279$\pm$.0046&.0167 &.0690\\ \hline 
LFR\cite{zemel2013learning} &.0182$\pm$.0211&.2201$\pm$.0318&.2496$\pm$.0044&.0044 &.0190\\ \hline 
\end{tabular}
\end{center}
\label{tab2}
\end{table*}


\begin{table*}[t!]
\def\arraystretch{1.3}
\small
\label{tab3}
\caption{Classification Performance 
on the Credit Card Data Set}
\vspace{-5pt}
\begin{center}
\begin{tabular}{l||c|c|c||c|c}\hline
\textbf{Method} & \bf Statistical 
Parity & \bf Normed Disparate & 
\bf Classifier Error & \bf Error Parity & \bf Error Disparate\\ \hline\hline
FRR\cite{calders2013controlling} & .0994$\pm$.0016 & .3109$\pm$.0186 
& .2340$\pm$.0058 & .0523&.1882 \\ \hline 
DFRR&.0118$\pm$.0006&.2038$\pm$.0627&.2283$\pm$.0062&.0250 &.1003 \\ \hline\hline
FKRR\cite{perez2017fair} &{.0079$\pm$.0011}&.1170$\pm$.0117&.2001$\pm$.0054&.0374 &.1643\\ \hline
DFKRR &.0085$\pm$.0015&.0957$\pm$.0286&.1823$\pm$.0092&.0306&.1151\\ \hline\hline 
FGR\cite{kamishima2012fairness}&.0779$\pm$.0571&.1283$\pm$.0987&.2412$\pm$.0469&.0253 &.0951 \\ \hline 
DFGR&.0494$\pm$.0601&.1221$\pm$.0890&.2244$\pm$.0382&.0105&.0442  \\ \hline\hline 
FPCA1\cite{samadi2018price}&.1716$\pm$.0149&.1458$\pm$.0234&.4025$\pm$.0382&.0941 &.2277 \\ \hline
FPCA2\cite{olfat2018convex}&.0981$\pm$.0164&.1307$\pm$.0193&.3224$\pm$.0045&.0663 &.1859 \\ \hline 
DFPCA&.0344$\pm$.0061&.1249$\pm$.0915&.2304$\pm$.0041&.0316 &.1230 \\ \hline\hline 
LFR\cite{zemel2013learning}&.0288$\pm$.0132&.1552$\pm$.0133&.2835$\pm$.0051&.0374&.1423 \\ \hline 
\end{tabular}
\end{center}
\label{tab3}
\end{table*}

\subsection{Other Analysis}

\begin{figure}[t!]
\centering
\includegraphics[width=.35\textwidth]{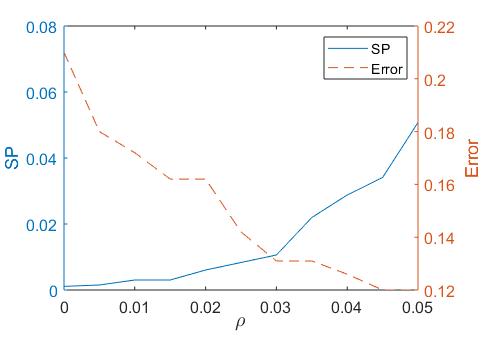}
\vspace{-5pt}
\caption{Classification Performance versus Threshold $\rho$}
\label{fig:2}
\end{figure}

\begin{figure}[t!]
\centering
\includegraphics[width=.3\textwidth]{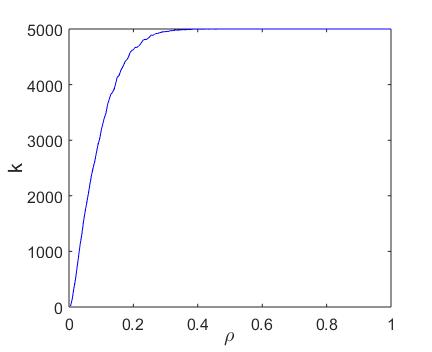}
\vspace{-5pt}
\caption{${k}$ versus ${\rho} $ (fixing ${\sigma=1}$)}
\label{fig:4}
\end{figure}

We first examined performance of the proposed 
distributed and private fair logistic regression 
on the Community Crime data set. The performance 
versus different $\rho$, averaged 
over 50 random trials and m = 5000, 
is shown in Figure \ref{fig:2}. 
We see as $\rho$ decreases, the classifier error 
increases and SP decreases. 
This means the model is fairer but 
less accurate, which is consistent with 
the implications of Theorems \ref{thm:finalfair} 
and \ref{thm_error}. (And considering that 
larger $\rho$ implies larger $k$, 
according to Lemma \ref{lemma_k} -- the implication 
of this lemma is verified in Figure \ref{fig:4}.) 

\begin{figure}[t!]
\centering
\subfigure[]{
\begin{minipage}[b]{0.22\textwidth}
\includegraphics[height=3.2cm,width=4.3cm]{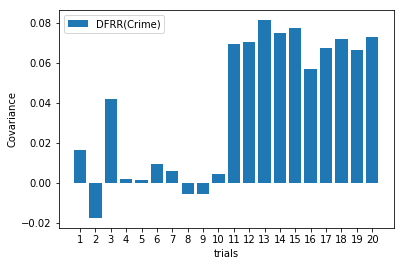} 
\end{minipage}
}
\subfigure[]{
\begin{minipage}[b]{0.22\textwidth}
\includegraphics[height=3.2cm,width=4.3cm]{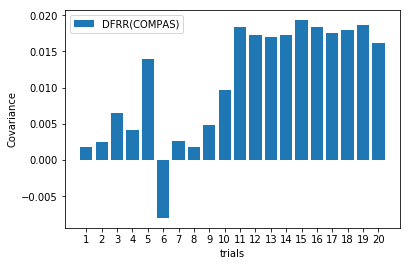} 
\end{minipage}
}
\caption{$\cov(f(x),s)$ of 20 random trials on the 
(a) Community Crime data set and (b) COMPAS data set}
\label{fig:6}
\end{figure}

Finally, we examined the PQD/PND assumption
in Theorem \ref{thm:finalfair}. 
Figure \ref{fig:6} shows $\cov(f(x),s)$ 
of DFRR over 20 random trials on two data sets. 
We see the covariance 
is positive in most cases, which implies 
$f(x)$ and $s$ are PQD/PND.

%% file: conclusion.tex
\section{Conclusion}

In this paper, we propose a distributed 
fair machine learning framework for protecting 
the privacy of demographic data. We propose a 
principled strategy to design private fair 
learners under this framework. 
We exemplify how to apply this strategy to 
redesign four non-private fair learners into 
private ones, and show our redesigns consistently 
outperform their non-private counterparts across 
three real-world data sets. Finally, 
we theoretically analyze the framework and 
prove its output models are both fair and accurate.

%% file: appendix.tex
\section{Appendix}

\subsection{Proof of Lemma \ref{lemma_fairF}}

We will prove that $f$ is fair 
if it is spanned by a set of fair hypotheses. 
Indeed, by the linear property of covariance, 
\begin{align}
\label{eq:app_l3a}
\begin{split}
\cov[f(x), s] 
& = 
\cov \left[ 
{\sum}_{t=1}^{k} \alpha_{t} 
h_{r_{t}}(x),\, s \right]\\ 
& = 
{\sum}_{t=1}^{k} \alpha_{t} 
\cov[ h_{r_{t}}(x),\, s]\\ 
&\leq ||\vec{\alpha}|| \cdot 
||\,\vec{\cov}[h_{r},\, s]\,||, 
\end{split}
\end{align}
where $\vec{cov}[h_{r},s] 
= [\, \cov[h_{r_{1}}(x),s], 
\ldots, \cov[h_{r_{k}}(x),s]\, ]$ 
and the last inequality is 
by the Cauchy–Schwarz inequality.
Further
\begin{equation}
\label{eq:app_l3b}
||\vec{\cov}[h_{r},s]||^{2} = 
\sum_{t=1}^{k} 
\cov[h_{r_{t}}(x), s]^{2} \leq 
\sum_{t=1}^{k} \rho^{2} = k \rho^{2},   
\end{equation}
where the inequality is based 
on the fact that returned 
hypotheses are $\rho$-fair. 
Combining (\ref{eq:app_l3a}) and 
(\ref{eq:app_l3b}) proves the lemma.

\subsection{Proof of Lemma \ref{lemma_k}}

First note the expected number of 
returned hypothesis is 
\begin{equation}
\label{theory_k3}
E [k] = m \cdot p( \cov(h(x),s) \leq \rho). 
\end{equation}

We will bound the right side probability. 
For convenience, write $\cov(h,s)$ for $\cov(h(x),s)$. 

The first result is a direct result 
of the Markov inequality. 

To prove the second result, we use 
the Chebyshevs inequality. It states 
that, over random $h$, 
\begin{equation}
\label{eq35}
\Pr \{ |\cov(h,s) - \text{E}[\cov(h,s)]| 
\geq \rho\}  
\leq \frac{\text{Var}[\cov(h,s)]}{\rho^2}.  
\end{equation}

We will refine (\ref{eq35}). 
We first show $\text{E}[\cov(h,s)]$ = $0$. 
This is true because $h$ is linear, 
i.e., ${h(x) = h^{T}x}$, so that 
\begin{align}
\begin{split}
\cov(h(x),s) &= E[h(x) s] - E[h(x)] E[s]\\
& = E[h^{T}x\, s]-E[h^{T}x]E[s]\\
& = h^{T} (E[x\, s]- E[x]E[s])\\
& = h^{T} \vec{\cov}(x,s), 
\end{split}
\end{align}
where $\vec{\cov}(x,s)$=$ 
E[x\, s]-E[x]E[s]$ is a constant vector. 
Then, taking expectation of $h$ on both 
sides, we have 
\begin{equation}
\label{eq:app_l5a}
E [\cov(h,s)] 
= E[h^{T} \vec{\cov}(x,s)] 
 = E[h]^{T} \vec{\cov}(x,s) = 0,  
\end{equation}
where the last inequality holds because 
$h$ is from a zero-mean normal distribution 
and thus $E[h]=0$.  

Next, we derive $\text{Var}[cov(h,s)]$.  
\begin{align}
\label{eq:app_l5b}
\begin{split}
Var[\cov(h,s)] = 
& Var[h^T \vec{\cov}(x,s)]\\
=& \vec{\cov}(x,s)^{T} \cdot 
Var[h] \cdot \vec{\cov}(x,s)\\ 
=& \vec{\cov}(x,s)^{T} \cdot 
\sigma^{2} I \cdot \vec{\cov}(x,s)\\ 
=& \sigma^2 ||\vec{\cov}(x,s)||^2.
\end{split}
\end{align}

Plugging (\ref{eq:app_l5a}) 
(\ref{eq:app_l5b}) back to 
(\ref{eq35}) proves the lemma. 

\subsection{Proof of Lemma \ref{thm_sp}}

Suppose $f(x)$ and $s$ are PQD or NQD random 
variables with bounded covariance. 
If $f$ is $\rho$-fair w.r.t. $s$, then 
\begin{equation}
\SP(f) \leq  \rho / s_{0} s_{1},    
\end{equation} 
where $s_{0} = p(s=0)$ and $s_{1} = p(s=1)$.

We will apply Lemma \ref{thm:lm1}. 
Recall $f(x), s \in \{ 0 , 1 \}$. The 
trick is to set $f(x) = 0$ and $s = 0$. 
Then, by Lemma \ref{thm:lm1}, 
\begin{equation}
\label{eq:app_l2a}
|H_{f,s}(0,0)| \leq 
\sup |H_{f,s}(0,0)| \leq |\cov(f,s)| \leq \rho.
\end{equation}

Now we refine $H_{f,s}(0,0)$. 
Write $f$ for $f(x)$. Note that 
\begin{align}
\begin{split}
H_{f,s}(0,0) 
& = F_{f,s}(0,0) - F_{f}(0) F_{s}(0) \\
& = p(f=0, s=0) - p(f=0) p(s=0). 
\end{split}
\end{align}

Plugging in $p(f=0) = p(f=0,s=0) + p(f=0,s=1)$ 
and rearranging terms, we have 
\begin{equation}
H_{f,s}(0,0) 
= s_{0} p(f=0, s=0) + s_{1} p(f=0,s=1), 
\end{equation}
where $s_{0} = p(s=0)$ and $s_{1} = p(s=1)$. 
Plugging this back to (\ref{eq:app_l2a}) and 
dividing both sides by $s_{0} s_{1}$,  we have 
\begin{equation}
|p(f=0, s=0) / s_{0} - p(f=0,s=1) /  s_{1}| 
\leq \rho / s_{1} s_{0}.      
\end{equation}
The left side is $\text{SP}(f | s)$. 
Thus the lemma is proved.

\subsection{Proof of Lemma \ref{lemma_distortion}}

The proof sketch is similar 
to \cite{arriaga2006algorithmic}. 
Note $||\tilde{x}||^{2} = ||H^{T} x||^{2} 
= {\sum}_{i=1}^{k} \langle h_{r_{i}}, x\rangle^{2}$.
Since each element of $h_{r_{i}}$ is from 
Gaussian, $\langle h_{r_{i}}, x\rangle$ is 
also from Gaussian and thus by definition $||\tilde{x}||^{2}$ 
is from a Chi-Squared distribution with $k$ 
degrees of freedom. Define a scaled variable 
$Z = \frac{k}{||x||^{2}} ||\tilde{x}||^{2}$; 
it is also from Chi-Squared with the following 
moment generation function
\begin{equation}
f(\lambda) = \mathbf{E}(e^{\lambda Z}) 
= \frac{e^{\frac{\bar{\eta} \lambda}{1-2\lambda}}}{(1-2\lambda)^{k/2}}, 
\end{equation}
where $\eta = \sum_{i=1}^{k} 
\langle h_{*}, x\rangle^{2} 
= k \langle h_{*}, x\rangle^{2}$. 
By the Markov inequality, 
for $0 \leq \lambda < 1/2$,
\begin{align}
\begin{split}
\Pr(Y \ge (1+c)k) 
&\leq \frac{e^{\frac{{\eta}\lambda}{1-2\lambda}} \cdot e^{-(1+c)k\lambda}}{(1-2\lambda)^{k/2}} \\ 
& \leq e^{\frac{{\eta}\lambda}{1-2\lambda} 
+ 2\lambda^{2}k - ck\lambda}\\ 
& = e^{\frac{{\eta} c}{4-2c} - \frac{c^{2}k}{8}},  
\end{split}
\end{align}
where the last equality is obtained 
by setting $\lambda$ = $c/4$. 

By similar argument 
(setting $\lambda$ = $c/2$), we have 
\begin{equation}
\Pr(Y \le (1-c)k) \leq 
e^{\frac{-{\eta} c}{2+2c}-\frac{c^{2}k}{4}}. 
\end{equation}
Combining the above two results via 
a union bound, we have
\begin{equation}
\Pr(|Y - k| \ge ck)  
\leq e^{\frac{{\eta} c}{4-2c} - \frac{c^{2}k}{8}} 
+ e^{\frac{-{\eta} c}{2+2c}-\frac{c^{2}k}{4}}
\leq g(x) e^{-\frac{c^{2}k}{8}},     
\end{equation}
where 
$g(x) = e^{\frac{k \langle h_{*}, 
x\rangle^{2}}{4/c - 2}} 
+ e^{\frac{- k \langle h_{*}, 
x\rangle^{2}}{2/c + 2}}$. 
The range of $c$ follows the range 
of $\lambda$. The lemma is proved. 

\subsection{Proof Sketch of 
Theorem \ref{thm_error}}

The original generalization error bound 
is developed using a distortion bound 
$\Pr\{ |\,||\tilde{x}||^{2} - ||x||^{2}\,|
\geq c ||x||^{2}\} \leq 2 e^{- 
\frac{c^{2} k }{8}}$, which assumes 
zero-mean distribution of projection 
vectors. Here, we use the new distortion 
bound in Lemma \ref{lemma_distortion}. 

Recall $x$ is an instance and $\tilde{x} = H^{T} x$ 
is its projection in a random space. If $f$ is linear, 
let $\tilde{f} = H^{T} f$ be its projection. 
If $||f|| = ||x|| = 1$, by similar arguments 
in \cite[Lemma 3.2]{garg2002generalization}, 
\begin{equation}
\Pr \{ \text{sign} \langle f, x\rangle 
\neq \text{sign}  \langle \tilde{f}, \tilde{x} 
\rangle \} \leq g(x) e^{-\frac{k \langle f, x\rangle^{2}}{8(2+|\langle f, x\rangle|)^{2}}},  
\end{equation}
where in $g(x)$ the coefficient is 
$c = |\langle f, x\rangle| / 
(|\langle f, x\rangle| + 2)$. 

Then, by similar arguments in 
\cite[Lemma 3.4]{garg2002generalization}, 
the classification error $\epsilon_{r}$ caused 
by random projection satisfies  
\begin{equation}
\label{proof_e1}
\epsilon_{r} \leq 
\frac{1}{n}\frac{1}{\delta} \sum_{i=1}^{n} 
g(x_{i}) e^{-\frac{k \langle f, x_{i}\rangle^{2}}{
8(2+|\langle f, x_{i}\rangle|)^{2}}},   
\end{equation}
with probability at least $1 - \delta$ over 
the randomness of $H$. 

Finally, by similar arguments in 
\cite[Theorem 3.1]{garg2002generalization} 
but with (\ref{proof_e1}) plugged in, 
with probability at least $1 - 4 \delta$, we have 
\begin{align}
\label{proof_sum}
\begin{split}
er(f) 
& \leq \hat{er}_{s}(f) + T + 
\epsilon_{r} + \epsilon_{r} \\ 
& \leq \hat{er}_{s}(f) + T + 
\frac{2}{n\delta} \sum_{i=1}^{2n} 
g(x_{i}) e^{-\frac{k \langle f, x_{i}\rangle^{2}}{
8(2+|\langle f, x_{i}\rangle|)^{2}}},  
\end{split}
\end{align}
where $T$ is a standard error 
bound from VC theory 
\begin{equation}
T = 2\sqrt{\frac{(k+1) \log(2en/(k+1))
+ \log(1/\delta)}{2n}}. 
\end{equation}

In (\ref{proof_sum}), there are $2n$ instances 
in the sum due to the introduction of a 
duplicated sample during proof. 
Replacing $2n$ with $n$ gives our theorem.